# Deep Reinforcement Learning for Optimizing Energy Consumption in Smart Grid Systems

Abeer Alsheikhi[1] ,Amirfarhad Farhadi[2], Azadeh Zamanifar[3]

[1,2] School of Computer Engineering, Iran university of Science and Technology, Tehran, Iran

[3] Department of Computer Engineering, SR.C, Islamic Azad university, Tehran, Iran

**Graphical Abstract**

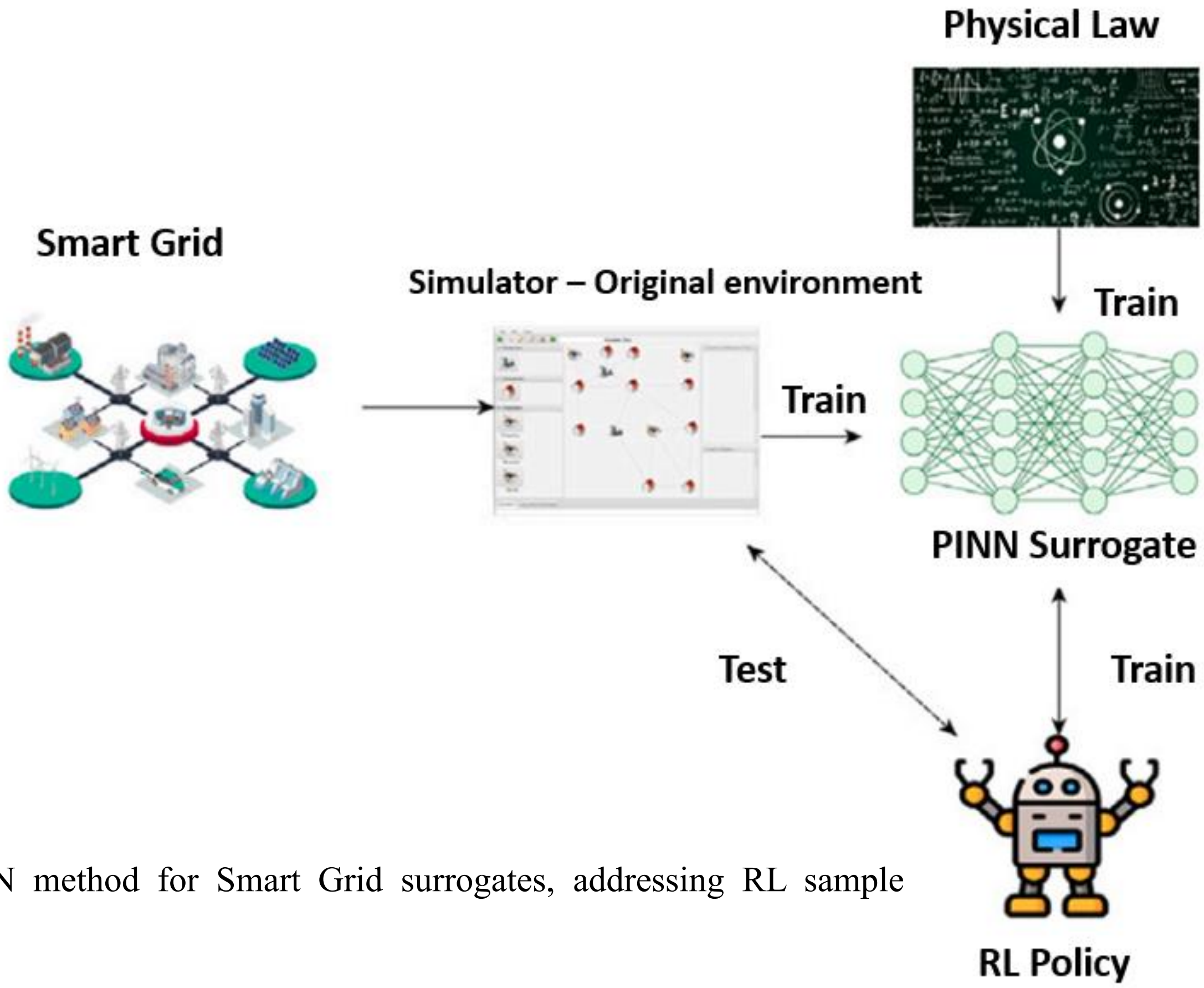

**Highlight**

- Novel PINN method for Smart Grid surrogates, addressing RL sample efficiency.
- PINN surrogates outperform data-driven models in RL training for Smart Grid management. PINN achieves 50 % faster RL policy training with comparable performance.
- Efficient RL training without costly simulator samples via physics-based surrogates.

## Abstract

The energy management in the context of smart grids is a kind of real-world problem complex by nature and due to the interrelatedness among diverse elements that are part of these scenarios. Although Reinforcement Learning (RL) has been proposed for solving Optimal Power Flow (OPF) problems, the requirement of iterative interaction with an environment can force it to employ computationally complex simulators, generating significant sample inefficiency.

In this study, these obstacles are overcome by the use of Physics-Informed Neural Networks (PINNs), which can replace conventional and expensive smart grid simulators. The RL policy learning is fine-tuned so that the convergent results be reached during a fraction of the time taken by the original environment. The PINN-based surrogate was compared to other benchmark Data-Driven surrogates. With the knowledge of the underlying physical laws, it was found that the PINN surrogate is the only approach considered in this context that can obtain a strong RL policy even without having samples from the true simulator.

The results show that using PINN surrogates can accelerate training by 50% as compared to RL training without surrogate. This method allows for fast generation of performance scores similar to the original simulator output.

**Keywords:** Smart grids control; Reinforcement learning; Physics-informed neural networks; Active network management; Optimal power flow; Surrogate models; Renewable energy

## 1.Introduction

The development of "smart grids" is now driving the modernization of our electricity networks. These cutting-edge systems are a direct reaction to the worldwide push to reduce pollution, more efficiently utilize energy and maintain consistent electricity supply through load adjustment. The European Union is leading this transition and has targeted that at least 43% of its energy should be generated from renewables by 2030 [1]. They are already doing so to some extent: green energy reached 32.1% in 2021 [2].

Renewable energy sources like wind and sun may not always be generating power when we need them, so today's society also requires a grid that can effectively prepare for and handle that variation. Such networks must be able to employ energy storage and demand side management tools in a controllable and intuitive manner, while remaining affordable and reliable for the general public [3]. This evolution has already gained significant traction in the EU with more than 13 countries rolling out smart meters to over 81% of homes. This transformation enables local energy communities, and flexible power markets [4], to further develop.

And the digital revolution, which has already given us smart meters, enables more sophisticated methods for this "demand response." Outside of energy systems, we have seen GenerativeAI used to automate administrative and workflow activities that are time-consuming in other high-stakes domains, thereby lightening the burden of operations while raising fairness, safety, and trust issues—suggesting a need for prudence when deploying AI in real-world decision support systems [54]. They do so by load-balance grid, decreasing the electricity usage at peak hours, and moving them to off-peak hours [5]. Yet, synthesizing digital models for these networks is a major challenge as we are now integrating more and more decentralized generation and storage, which has its counterpart at the Big Data level.

To determine how best to operate such modern grids, scientists are turning to simulators on which they can try out data-driven Machine Learning (ML) algorithms. Classical mathematical equations have been the standard, but augmenting them with ML models like decision trees or neural networks has emerged as a powerful tool for generalizing across such diverse phenomena from individual lithium-ion batteries [7,8] to entire national power systems [9,10].

Reinforcement Learning (RL) has the potential for success in this domain as it provides a "trial-and-error" reward mechanism to identify best possible techniques for managing energy storage and power distribution [11,12]). Artificial Neural Networks as Other Predictive Modelling Tool When combined with RL, ANNs or other predictive modelling tools allow researchers to smartly design policies for consumer demand management [13,14]. The challenge is that RL requires constant engagement in the system to learn from it. If we attempted to teach such AI agents on an actual, operational electricity network, they could make a mistake that causes the grid to black out or equipment damage. This is why safe simulation environments are critically important for the training.

More generally recent work on Generative AI has highlighted that while such models can be very flexible the deployment of them at scale in safety-critical applications must have frameworks to govern privacy, security and ethical risks [55].

In our work we developed a realistic grid model using Gym-ANN. The benefit of this allows for the incorporation of real-world physical constraints, and various types of physical hardware such as local generators and storage devices enabling training of RL policies that are fully deployable into practice [15].

The disadvantage is that the simulations are very low as one deals with complex patterns and high-resolution grids [16]. To circumvent this bottleneck, we suggest a novel methodology: Physics-Informed Neural Network (PINN). This serves as a high speed ''stand-in'' for the original simulator, yielding identical results within as small fraction of the time.

With this alternative model, we are able to train a RL policy without the actual grid environment. By this strategy, the policy can achieve the same near-optimal performance level as the original system but at much less time. Our findings stress the importance of taking into account the true physical laws of a system when constructing a model. We demonstrated it by comparing our approach with other modern "data-only" models: our method works unlike theirs thanks to its underlying physics and not just raw data."

We show that by employing these physics-based substitutes (or PINNs), it is possible to achieve a 50% reduction in the training duration as compared with directly training within the real simulation. This means we can get the results of equal quality in half the time. This advance is one of the most significant steps toward smart grids of the future, because it will allow us to much more easily control these very high-voltage systems and be able to integrate new technologies like solar energy in with the older infrastructure that we've always used.

To tackle the computational challenges and sample inefficiency associated with reinforcement learning-based optimal power flow solutions in smart grids, this study introduces a physics-informed surrogate modeling framework. A surrogate model based on Physics-Informed Neural Networks (PINNs) is developed to replicate smart grid dynamics while explicitly incorporating physical constraints derived from power system laws and the operational limits of devices. Specifically, the Karush-Kuhn-Tucker (KKT) conditions are integrated into the surrogate formulation to ensure the feasibility of generator and energy storage operations. This approach allows reinforcement learning policies to be trained without direct interaction with the original simulator, significantly reducing computational costs. A systematic comparison with state-of-the-art data-driven surrogate models shows that physics-informed surrogates offer enhanced robustness and generalization, particularly in regions of the state space that are either poorly sampled or completely unseen. Quantitative results indicate that the proposed method achieves comparable control performance while cutting reinforcement learning training time by approximately 50%, underscoring its potential for scalable and reliable smart grid control.

## 2. Background

### 2.1. Smart grid: Optimal power flow and control systems

Conventional attempts for solving OPF problems adopted numerically based approaches, most were based on the use of Newton-Raphson method and some other mathematical programs such as recursive-based programming, Lagrange multipliers and linear approach [18–20]. Traditional dispatch models concerning power systems grounded on OPF techniques seek the optimal operation of generation resources to supply grid demand in a cost-efficient way, whilst meeting system stability and some relevant technical limitations related to its elements [17].

The transition of the grid paradigm, with increasing penetration of renewable generation and distributed generation (DG) systems, and also inclusion of energy storage systems (ESSs), in addition to demand response requirements smart grid dispatch are also faced as new challenges for tackling efficiently OPF problems. Such complications result from complex constraints and dependencies on stochastic variables such as weather or state of charge of batteries [20,21].

Other utilities have simplified the OPF problem, from an AC perspective, to speed execution time by going DC in an attempt to simplify computation.

However, these simplifications caused sub-optimality in many cases and led to safety issues as well as non-realistic controllers [17]. On the other hand, data-centric machine learning methods provide a promising potential for computational burden reduction and system stability improvement. This requires a transition from online optimization methods to offline training with large amounts of historical or simulated data [17].

## 2.2. Surrogate models

With the increasing volume of information, efficient data reduction, such as summarization methods, has even been investigated to countermeasure computational and informational overloading, affirming the general motivation for compact representations and approximations in modeling complex systems [57]. Operating a modern smart grid is an extraordinary technical challenge, in part because the actual physical system is too complex to simulate with detailed models that can only be solved on supercomputing clusters. One way around this problem is with “surrogate models.” These are simple proxies for the actual system, but they return results far faster by making estimates about how the grid will operate [22].

For those trying to solve an optimal power flow (OPF), these models are orders of magnitude faster than old-school math formulas. These can more generally be grouped into two categories: Analytical (ASMs) and Learned (LSMs). Parallel concepts in the use of surrogate-assisted optimization are seen in other complex operational settings, such as supply chain management, for which adaptive neuro-fuzzy inference systems (ANFIS) and blockchain-based solutions have been investigated to enhance robustness and transparency within uncertain and nonlinear constraints [56]. For instance, some studies on the analysis and computation used the Kriging method to address the randomness of wind farms [23]. Others use it as a way to improve the "trial-and-error" math tools, such as increasing the speed and accuracy of Genetic Algorithms [24].

On the “Learned” side, there’s an even more unconventional approach. Others studied the possibility of getting rid from all data and use machine learning techniques in order to predict voltage levels in large networks consisting of more than 8,500 nodes [25]. These intelligent models can work twice as quickly as conventional methods such as Monte Carlo simulations.

Of course, there is a catch. To achieve that speed, you sometimes have to trade little accuracy. It also requires an enormous amount of time and effort to “train” these models in the first place, especially if you want them to obey rigid physical constraints. But the trade-off is mostly worth it, most experts believe. When the model is set up, it responds to nearly all requests at near-instantaneous speed, allowing for constant re-evaluation of the grid state.

## 2.3. Physics-informed machine learning

Data-driven surrogates, such as decision trees, random forests, XGBoost, and deep neural networks, may perform well within the training data distribution but struggle with out-of-distribution scenarios or underrepresented regions. PINNs offer an innovative approach that addresses many limitations of conventional surrogates [26]. Unlike traditional data-driven models, PINNs integrate physical equations, such as ordinary differential equations or partial differential equations, directly into the training loss function to ensure that the model’s output remains consistent with fundamental physical laws, leading to improved robustness in poorly sampled regions of the state space compared to purely data-driven models. This makes them particularly suitable for modeling systems with high complexity and variability, such as energy networks with dynamic load profiles and renewable energy generation.

For example, PINNs have been used to model lithium batteries by including the governing PDEs of electrochemical processes of Li-ion batteries [27], or mathematical models for Li-ion batteries such as the Single Particle Model [28], or the pseudo-2D model [29]. PINNs were also used to model parts of the behavior of hydrogen electrifiers, such as the temperature fluctuation of the membranes [30], or the voltage losses during the electrification [31].

Regarding smart grids, in [32,33], PINNs were developed, including active and reactive power balance equations, to solve AC-OPF problems, and in all study cases shown, PINNs were more accurate than other conventional Artificial Neural Network (ANN) schemes. The advantages brought by PINNs can also be applied to solutions beyond classification or forecasting, such as Reinforcement Learning, which is a strategy widely used in energy systems due to its reward-focused approach, as explained in the introduction. Moreover, model-free RL tends to struggle with training efficiency, often requiring excessive computing time to achieve a satisfactory reward value, and it also tends to violate physical constraints [34]. As such, coupling PINNs with RL algorithms can increase training performance and has drawn attention in multiple complex environments.

An example of which includes a physics-informed Deep Dyna Q RL for the control of building heating systems, in which PINNs were used for the planning phase of Dyna Q, which objective function combined a regression loss and a physics-based loss with the RC heating model [35]. Focusing more on the field of energy, PINNs were also used with PPO-based RL to extract parameters of PV, wind, and energy storage systems connected to a local grid, so that they could be used to accurately simulate and forecast the behavior of such grid and use the parameters to control and preserve the stability of the grid with RL [36].

**2.4. Reinforcement learning solution of the optimal power flow problem**

Reinforcement Learning (RL) is the methodology adopted in this study. In this configuration, an AI "agent" is trained on how to operate a smart grid through repeated interactions with it. The agent gets feedback — something like a grade report — on how well its actions did, and it wants to maximize this feedback over time. There are many researchers advocating RL use in power flow problems [37,38], however this is burdened with a fundamental challenge: complexity. Smart grids have too many moving parts, so the agent must train in a vast number of different situations. This results in a "curse of dimensionality", because it requires too many data to fill up each bucket so that the training process can become slow or not work at all [39].

It is hard to find a sufficiently realistic simulation for such training. A very known framework (like) is Grid2Op for the most detailed models of transmission system including power lines and generators constraints [40]. For instance, some researches which employ an RL technique named "Soft Actor-Critic"(SAC) in Grid2Op to explore the optimal approach of power dispatch aims to reduce costs and make full use of renewable energy, are adopted [41].

Another, network-distribution-centric tool is Gym-ANN [15]. Similar resilience issues also arise in distributed cyber-physical networks, such as wireless sensor and actor networks, where connectivity restoration under node failures is concerned with a fast, distributed recovery mechanism that also becomes relevant to reliability considerations for power system operations over a smart grid [58]. This model encompasses numerous smart grid components, such as flexible batteries and distributed energy generation. It gives scientists a way to design specific goals — such as maintaining an embedded battery's "State of Charge" (SoC) — that also respect the physical laws of the grid.

Unfortunately, attempting to bring slow physics simulations into a RL framework involving large (MB-scale) and frequent data reads here is also prohibitively inefficient. To ameliorate this problem, some have proposed the use of Machine Learning "prostitutes" (proxies) as a substitute to the true simulation [42–43]. Our work shows that even if these surrogates aren't perfect replicas, provided they adhere to some basic physical laws, they can still be highly effective. We think the correct answer is a hybrid that combines data-driven learning with basic physical understanding.

That is the inspiration for this paper. Using Physics-Informed Neural Networks (PINNs) trained with real physical laws, we demonstrate that it is possible to create RL systems able to learn signifi- cantly faster and in a more reliable manner than ever before. Our study builds upon the work of Cestero et al. [44]. Though those authors were able to construct well-trained surrogate models, they reported a significant shortcoming: their models didn't actually "understand" the physics of the system. Since those models considered only raw data, they were not suitable to train an AI for controlling a real-world system. We address this problem by creating a sur- rogate for the Gym-ANN environment similar to the one we developed herein, and compare its output to that of theirs in order to establish how much our new methodology is able to contribute.

## 3.Methods

In this paper we propose to solve the OPF problem by using RL to learn a working policy which can operate in a smart grid good enough on order is avoid penalties and reduce the energy loss of the grid. For that purpose we also use a smart grid simulation environment as our RL environment: the ANM6-Easy [15].

While the rendering is true to life in terms of grid bouncing, expensive iterates of mathematics are required to solve the equations that this environment needs. This renders the environment infeasible for larger topologies, so that scalability of this (RL based) solution to solve the OPF problem of a smart grid is limited. To address this problem, we propose to employ surrogates of the environment. However, as discussed by Cestero et al. [44], however using data-driven techniques to construct surrogates, even though are robust in representing the state transitions, they present good metrics $R^2$, sometimes is not enough to accurately represent the physics of state transitions. To this end, we suggest to leverage PINNs as an emulator and compare it with other data-driven approaches.

The architecture of our approach is depicted in Fig. 1. We consider two deferent ways of training the surrogate from its original environment: data-driven and PINN-based. For the PINN method, we take the physical laws from the original environments to derive them, and train a surrogate environment with no use of the original environment as detailed in [45]. On the other hand we employ some of the latest methods for constructing data-driven models (XGBoost, Decision Trees, etc.). We note that all the data-driven algorithms learn a training data set based on the original environment. By the methods of Cestero et al. [44], we have two types of datasets: a generative dataset, where we procure transitions from the environment not interacting among each other by computing the intermediate actions of a random sample across the state space of the environment; and an agent-based dataset, in which data was gathered using a random agent mimicking realistic trajectories based on interpolated action sequences (performed with higher-level actions). Then we train all data-driven models using these two datasets individually.

We then train an RL policy over the surrogate environments on each of these trained models via PPO [46]. At training we evaluate the policy after every update using one episode of the original environment with the in-training policy. In the end, we save the score of the episode, and use it to analyze the policy's evolution learning in each surrogate method.

All equations presented in this paper are based on the formulation in [REF]

### 3.1. Reinforcement learning training regimes for surrogate models

In this work, we use a surrogate model as the environment of the RL algorithms. An algorithmic template was introduced to bridge the connection among the predictive model and RL logic. This framework was implemented in compliance with the interface of the gym [47] and it enabled us to interact with the PPO RL training algorithm [46] from Stable-baselines3 library [48], by fixing all default hyperparameter values.

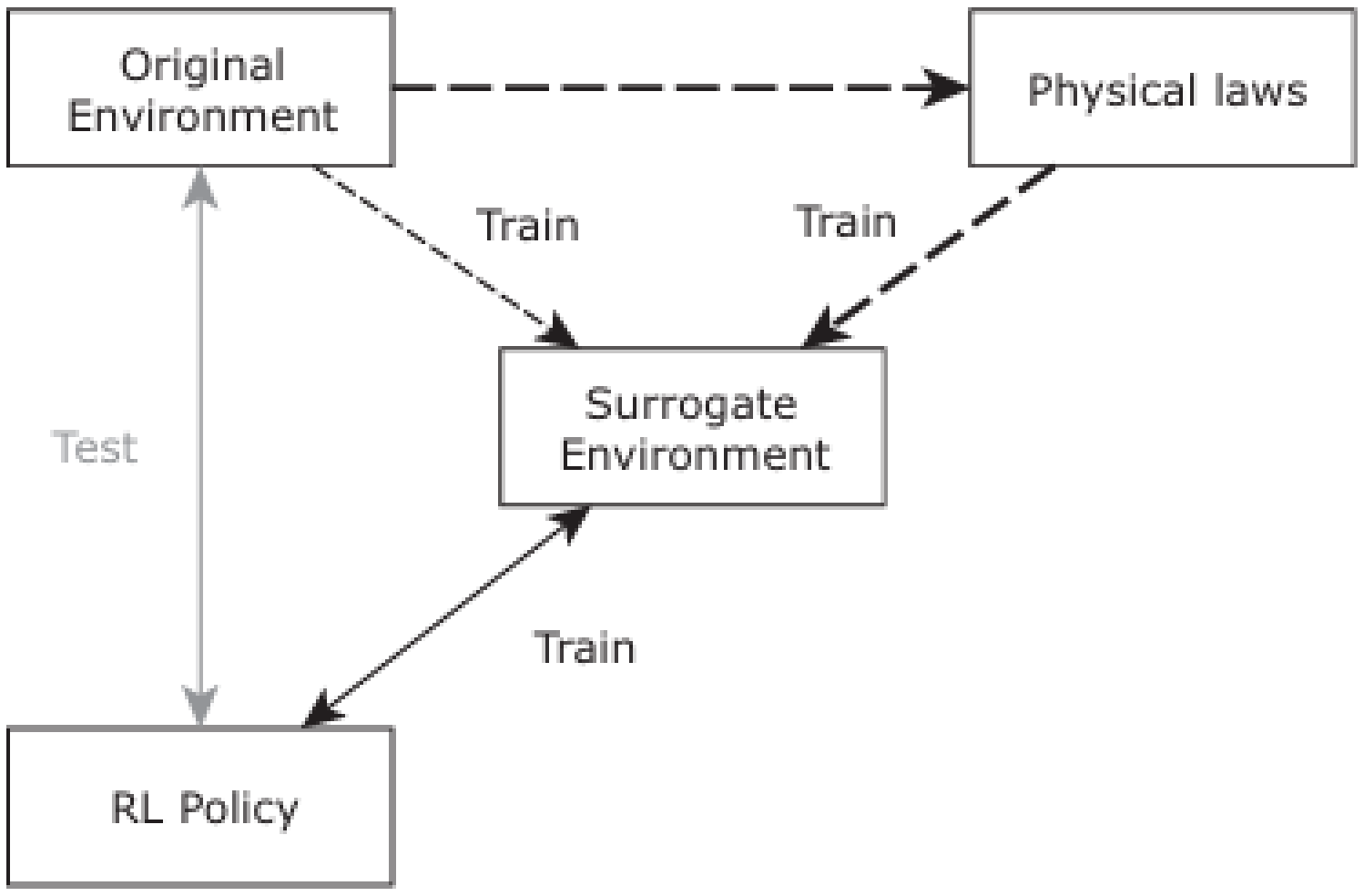

*Figure 1 The two different ways to create the surrogate environment – PINN method and data-driven approach – are shown here. The PINN training flow is shown with bold dashed lines and the data-driven process are presented with thin dashed lines. This surrogate environment is then used to train the RL policy. So that the progress measurement would not drift, policy is validated from time to time in the original environment with each update.*

The activities in the software framework are performed as follows:

- Initial state selection: The initial state of the agent is uniformly sampled from the original environment s space. Whenever the environment receives a reset signal, the initial state is sampled uniformly over all possible states.
- State transition computations: Transitions between states are executed by the surrogate model. The internal representation is then processed with the current state and agent's action to form a single input vector. This data is then used by the surrogate model to predict how the next state and a reward will change.
- Terminal states: The end of an environment episode is denoted by a “done” flag. The binary classifier based on gradient boosting is used, as the surrogate model cannot discern between solvable and unsolvable physical equations [49]. We use this classifier to recognize the terminal state with 99% accuracy.

**3.2. Parallelization of the surrogate environment**

We leverage the inherent parallelism in ANN based surrogates that derives from linear algebra. This is very different from the initial setting where one use sequential, iterative algorithms [25], such as Newton-Raphson and which prohibits again parallel work. With the surrogate being inherently parallel, the training of RL policies learns much faster as environment samples can be collected orders of magnitude faster.

Although (as in this study) some frameworks do not inherently facilitate parallelization, one way to overcome this is to use an ANN-based surrogate. A custom wrapper is developed to interface with the proposed framework and parallelized. This configuration brings in two specific structural parameters:

**n_envs:** This quantifies the number of parallel environments the RL algorithm operates over. Larger number of environments may lead to faster policy convergence, but at the cost of computation. In the Stable-baselines3 implementation of PPO, samples are collected in a so-called rollout_buffer. The size of this buffer is calculated as:

$$rollout_{buffer_{size}} = n_{envs} \cdot buffer_{size} \dots\dots\dots\dots..1$$

With the inclusion of more environments, this overall buffer size will get larger and at some point potentially slow down training due to the sheer size of your dataset.

**buffer_size:** When experimenting, we saw that how often you update your policy is more critical to learning than the total size of seen data. If the policy is updated often (with a smaller buffer_size) huge gains can be seen in performance early on. This parameter determines how many steps to forward before doing an update. So, after every parallel environment completes the steps set by buffer_size, The rollout buffer is rows full and we are ready to update the policy.

This is done by vectorizing state transitions. This enables the surrogate model to process transitions with all parallel environments at once using the following function:

$$f: (S_t, A_t) \rightarrow (S_{t+1}, r_t) \dots\dots\dots\dots\dots\dots\dots 2$$

Into this equation, $A_t$, and $r_t$ are the (selected from all n_envs ways) matrices and vectors of collected data.

**4. Application on smart grids**

The ANM6 was utilized in this investigation. This structure resembles an integrated smart grid with three separate buses. One bus serves a residential load with photovoltaic (PV) generation, another supports an industrial load paired with a wind generator, and the third connects an electric vehicle (EV) charging station to an energy storage system (ESS). All components are powered by a slack generator, which acts as the interface with the wider transmission network.

The key objective in this setting is that of finding a policy to minimize the total energy wastage. This must imply a reduction in partial line losses and partial transformer loss, avoiding energy wasting, due to network overload and also to reduce curtailment of RES. These goals are achieved by controlling the ESS power flow and curtailing PV and wind generation, as necessary. The excess or deficit energy demand is met by the slack generator.

A strong constraint is represented by the lines feeding industrial load and EV charging park, as these lines have lower available power than the possible peak demand of both loads.

This reflects a typical recent issue in distribution networks, where the increasing penetration of electric vehicles often leads to expensive investments required for infrastructure upgradings to accommodate higher peak demands [50]. Yet it is possible to avoid these costs by actively controlling the local renewable sources and storage systems, leading to better generation costs and network efficiency.

### 4.1. ANM6-easy environment

The tis thesis develops Reinforcement Learning environments to optimize the distribution grid control, employed using the Gym-ANM simulation environment. In this context, the use case ANM6-Easy is intended to model a variety of management tasks on a small distribution network [15].

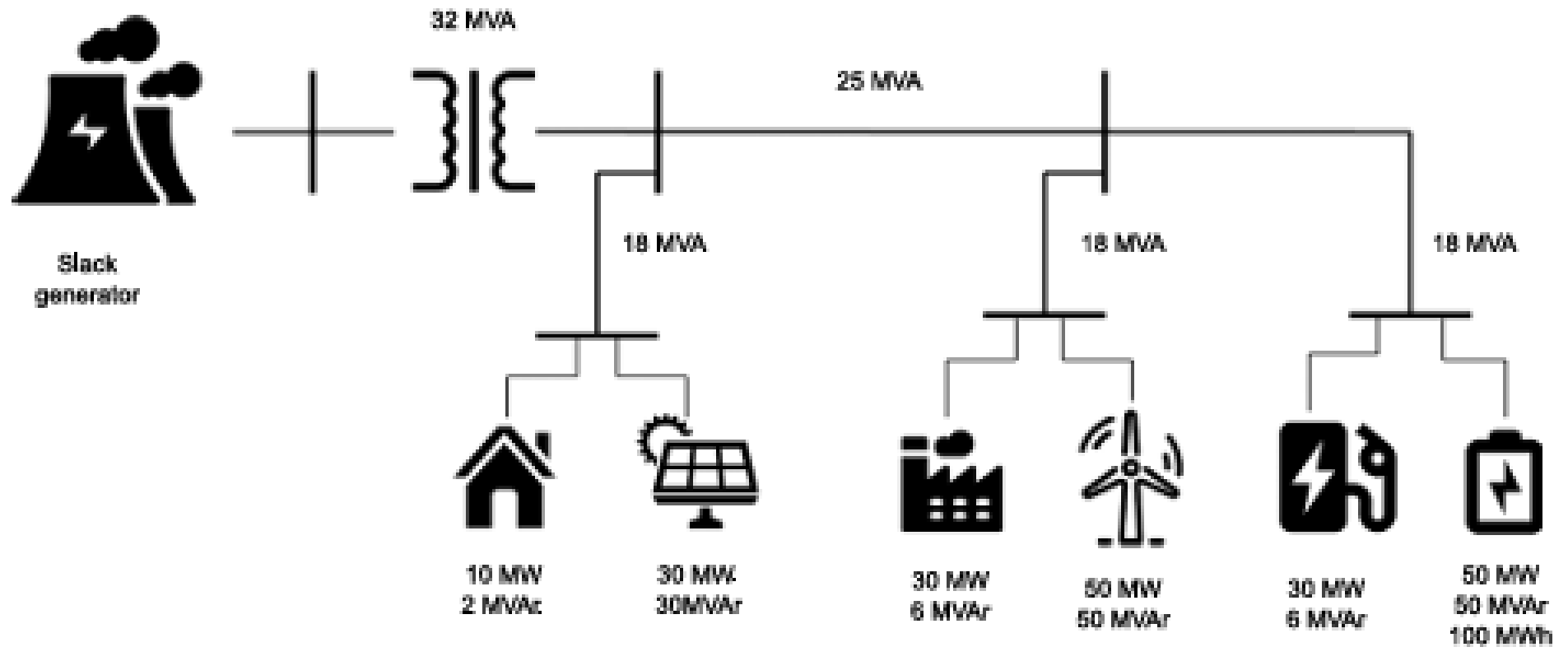

*Figure 2 Layout of the ANM6-Easy environment. The illustrated power system consists of a slack generator and two renewable generation units: wind farm and photovoltaic plant. The grid also contains three different passive loads including industrial, residential and EV charging station along with an individual energy storage unit.*

Tab.2ANM6-Easy model is composed of seven devices linked across six buses to take into account the distribution network. The devices in D are grouped into two meta-device groups: three generators (one is a slack generator) referred to as $D_G$,, and three loads, which are referred to as $D_L$,.

Further, there is a distributed energy source (ES) acting as an energy storing in a network. The slack generator is connected to its own bus (index 1) in order to balance power flows and to keep a voltage reference of $1p.u^0$All parts such as devices, buses and branches are characterized with physical properties/constraints [15].

Going to focus this study, for the sake of simplicity, an environment which uses deterministic. The load demand and the peak generation capacity is both represented by two deterministic, 24 hours' time series that carry on from one day to the other. The existence of such known patterns is postulated throughout the chapter—a property termed the "future-awareness hypotheses." The simulation time is discrete and expressed as $\Delta t = 15$min.

This future-awareness assumption enables to eliminate complexities in prediction errors. This enables a quantitative assessment of the PINN surrogate ability to fit well physical grid dynamics. Therefore, any policy performance or training speed gains can be justified in value of the quality of the surrogate model used, which we believe is the main merit of this paper.

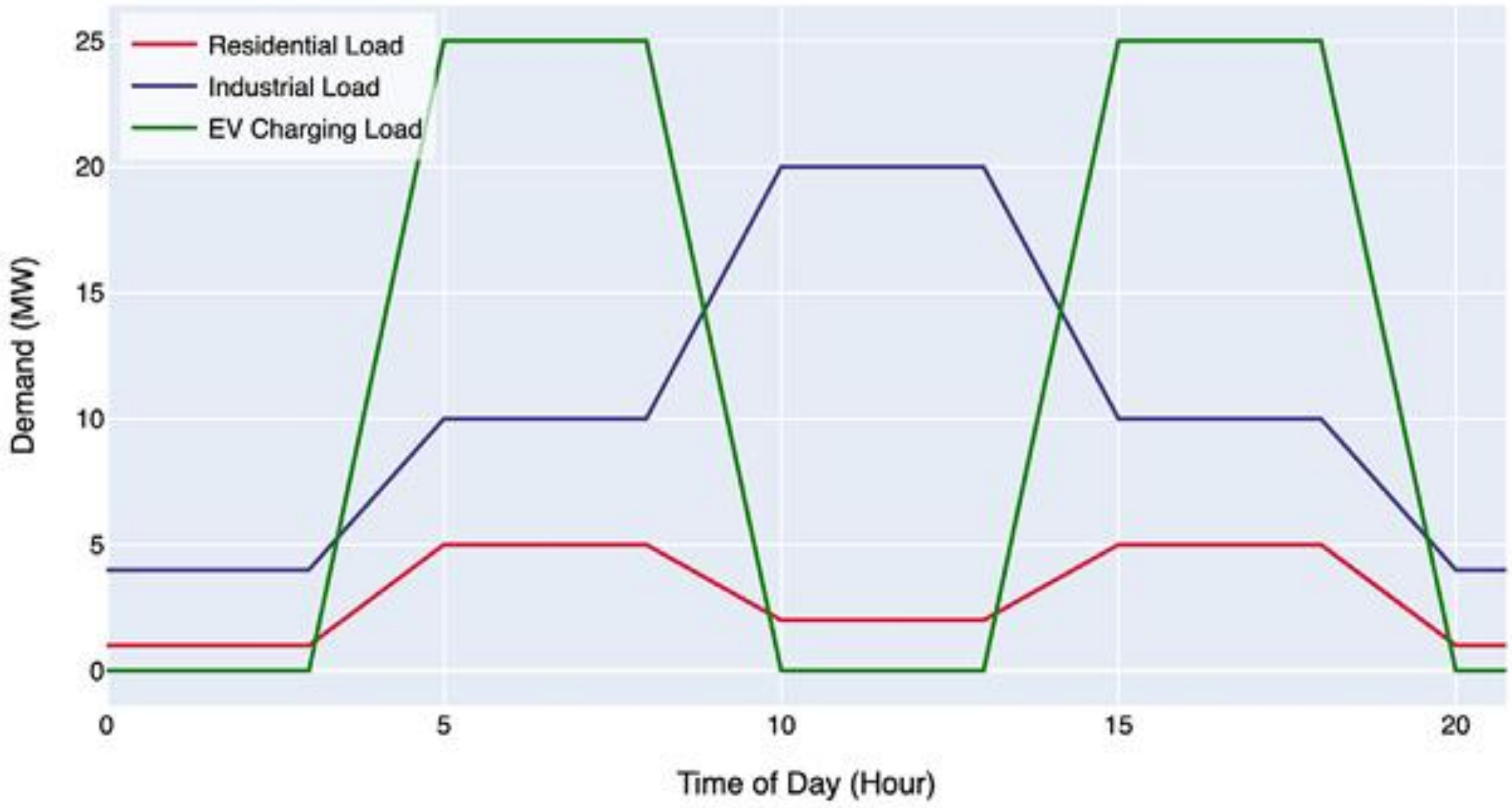

*Figure 3 Daily Load Demand Profiles*

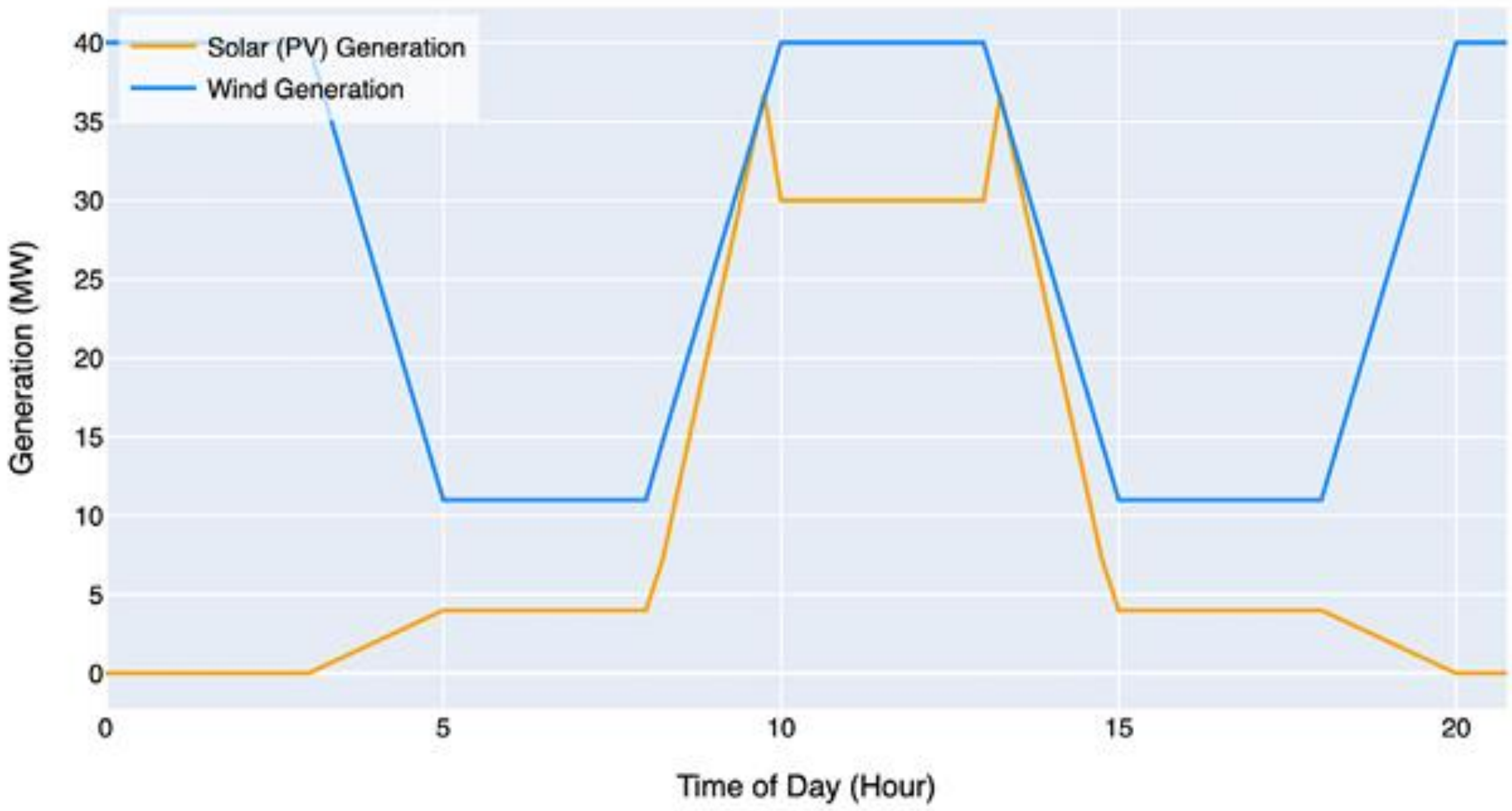

*Figure 4 Daily Potential Generation Profiles*

This arrangement also creates a best-case performance; it describes the best one can do with perfect forecasts. Such an assumption is of course a must for the present study, but the framework allows for modularity so that implementation in practice can be done. In a real-world scenario, the given time series would be replaced with data from a forecasting module and we would train the RL agent in an end-to-end PINN surrogate using forecast, uncertain values.

Production environment, the deterministic time series data would be replaced by the output of a dedicated forecasting module. The RL agent would then be trained within the PINN surrogate environment, but fed these uncertain, predicted values.

The state of the system can be represented for each step of time by the following equation.

$$\mathbf{s}_t = \left[\{P_{dt,t}\}_{d\in\mathcal{D}}, \{Q_{dt,t}\}_{d\in\mathcal{D}}, \mathrm{SoC}_t, \{P_{g,t}^{(\max)}\}_{g\in\mathcal{D}_G-\{g^{\mathrm{slack}}\},}, \mathrm{aux}_t\right] \ldots\ldots 3$$

Where:

- $P_{d,t}$ and $Q_{d,t}$ : The active and reactive power responsive at time t for all device
- $\mathrm{SoC}_t$ : Battery state of the charge.

  $P_{\mathrm{g},t}^{(\max)}$ : Maximum capacity produced by the generator g at time t ($g \in \mathcal{D}_G - \{g^{\mathrm{slack}}\}$)

- Aux: extra time component to preserve the Markov property of this process. It is used as an index, ranging from 0 to $95(24, \mathrm{h}/\Delta t)$, and facilitates obtaining the demand and the generation capacity for the next interval from pre-defined time series.
- Options for setting control variables at different network elements are captured by the action vector $\mathbf{a}_t$ at time step t. The particular form of $\mathbf{a}_t$ is set as:

$$\mathbf{a}_t = \left[\{a_{P_{8,t}}\}_{g\in\mathcal{D}_G-\{g^{\mathrm{slack}}\},}, \{a_{Q_{g,t}}\}_{g\in\mathcal{D}_G-\{g^{\mathrm{slack}}\},}, a_{P_{\mathrm{DES},t}}, a_{Q_{\mathrm{DES},t}}\right] \ldots\ldots 4$$

where:

- $a_{P,}$ and $a_{Q_{\mathrm{g},}}$ represent the active and reactive power setpoints, respectively,

  $D_G$ : Group of generators and $g^{\mathrm{slack}}$ : Slack generator

- $a_{P_{\mathrm{DES},t}}$ and $a_{Q_{\mathrm{DES},t}}$ denote the active and reactive power modifications for the Distributed Energy Systems (DES), which can together inject and remove power to equilibrium the system

### 4.1.1. Transition dynamics

Given the known future demand $\{P_{d,t+1}\}_{d\in D_L}$, and power factor pf of each load, reactive power of loads $\{Q_{d,t+1}\}_{d\in D_L}$ is calculated as:

$$Q_{d,t+1} = P_{d,t+1}\tan(\arccos(pf))\ \forall d \in \mathcal{D}_{L}.\ \dots\dots\dots\dots 5$$

For generators, constraints define feasible active and reactive power ranges, ensuring each device operates safely. The setpoints for active $a_{P_{g,t}}$ and reactive $a_{Q_{g,t}}$ power is mapped to the feasible set $\mathcal{R}_{g,t,\,a}$ convex polytope defined by generators physical parameters, network conditions, and external variables:

$$\begin{aligned}\mathcal{R}_{g,t.} = \{\,(P,Q) \in \mathbb{R}^2 \mid \underline{P}_{g.} \le P \le P_{g,t.}^{(\max)},\\ \underline{Q}_g \le Q \le \bar{Q}_g,\\ Q \le \tau_g^{(1)}P + \rho_g^{(1)},\\ Q \ge \tau_g^{(2)}P + \rho_g^{(2)}\},\end{aligned} \qquad \dots\dots\dots\dots\dots 6$$

The generator's reactive power injection flexibility constraints (line constraints) applied on a linear in system using known constants$\tau_g^{(1)}, \rho_g^{(1)}, \tau_g^{(2)}$, and $\rho_g^{(2)}$, for instance when the generator is close to the maximum of its active Power Capability Curve are presented.

This mapping is posed as a convex optimization problem. The most important criterion for this operation is to find the feasible point in $\mathcal{R}_{g,t}$ that distance from chosen setpoints $\left(a_{P_{g,i}}, a_{Q_{g,i}}\right)$ is minimized.

$$\left(P_{g,t+1}, Q_{g,t+1}\right) = \underset{(P.Q)\in D_{g,t}\subseteq\mathbb{R}^2}{\operatorname{argmin}} \| \left(a_{P_{g,t},a_{Q_{g,t}}}\right) - .(P.,Q)\|.. \ \dots\dots\dots\dots\dots\dots\dots .7$$

Similar approach is used for the DES unit plus supplementary limitations based on the SoC are introduced. The feasible operational set $\mathcal{R}_{\mathrm{DES}\,,t}$ is given by:

$$
\begin{aligned}
\mathcal{R}_{\mathrm{DES},t} = \Big\{ & (P,Q) \in \mathbb{R}^2 \mid P_d \le P \le P_d^{(\max)}, \\
& \underline{Q}_d \le Q \le \bar{Q}_d, \\
& Q \le \tau_d^{(1)} P + \rho_d^{(1)}, \\
& Q \ge \tau_d^{(2)} P + \rho_d^{(2)} \\
& Q \le \tau_d^{(3)} P + \rho_d^{(3)}, \\
& Q \ge \tau_d^{(4)} P + \rho_d^{(4)}, \\
& P \ge \frac{1}{\eta \Delta t}\left(\mathrm{SoC}_{t-1} - \overline{\mathrm{SoC}}\right) \\
& P \le \frac{\eta}{\Delta t}\left(\mathrm{SoC}_{t-1} - \underline{\mathrm{SoC}}\right) \Big\},
\end{aligned}
\qquad \ldots\ldots\ldots\ldots\ldots\ldots.8
$$

In this expression, $\eta$ serves for both charge and discharge efficiency. Additionally, the power constraints are expressed as:

$$
\left(P_{\mathrm{DES},t+1}, Q_{\mathrm{DES},t+1}\right) = \underset{(P,Q)\in\mathcal{D}_{\mathrm{DES},t+1}\subseteq\mathbb{R}^2}{\operatorname{argmin}} \left\| \left(a_{P_{\mathrm{DES},t}}, a_{Q_{\mathrm{DES},t}}\right) - (P,Q) \right\| \ldots\ldots\ldots.9
$$

The state of charge (SoC) for each DES unit is then updated as follows, based on the following equation:

$$
\mathrm{SoC}_{t+1} = \begin{cases} \mathrm{SoC}_t - \eta \Delta t P_{\mathrm{DES},t+1} & \text{if } P_{\mathrm{DES},t+1} \le 0 \\ \mathrm{SoC}_t - \dfrac{\Delta t}{\eta} P_{\mathrm{DES},t+1} & \text{otherwise.} \end{cases} \quad \ldots\ldots 10
$$

After that, the active and reactive power values of each unit (excluding the slack generator) are found and added up for each bus to be the net injections at that bus which are denoted as $\left\{P_{i,t+1}^{(\mathrm{bus})}\right\}_{i=2}^{6}$ and $\left\{Q_{i,t+1}^{(\mathrm{bus})}\right\}_{i=2}^{6}$. . And the resulting equations for the power flow problem can be solved using these injections along with the slack bus voltage definition and branch admittance matrix $Y_{ik} = G_{ik} + jB_{ik}$, as previously mentioned.

After calculating the active and reactive power of each element (the slack generator is excluded), it is used to group and sum the numbers according to own bus where we obtain the net injections $P_{i,t+1}^{(\mathrm{bus})}$. And $Q_{i,t+1.}^{(\mathrm{bus})}$ .By applying these injections with the known slack bus voltage and the branch admittance matrix $Y_{ik} = G_{ik} + jB_{ik}$,, the power flow equations yield:

$$P_{i,t+1}^{(\text{bus})} = \sum_{k=1}^{6} |V_{i,t+1}||V_{k,t+1}|(G_{ik,}\cos\,\theta_{ik,t+1.} + B_{ik}\sin\,\theta_{ik,t+1}) \ldots\ldots..11$$

$$Q_{i,t+1.}^{(\text{bus})} = \sum_{k=1}^{6} |V_{i,t+1}||V_{k,t+1}|(G_{ik.}\sin\,\theta_{ik,t+1.} + B_{ik}\cos\,\theta_{ik,t+1}) \ldots\ldots.12$$

where $P_{i,t+1}^{(\text{bus})}$and $Q_{i,t+1}^{(\text{bus})}$denote the active and reactive power injections at bus $i$at time step $t+1$; $|\ V_{i,t+1}\ |$and $|\ V_{k,t+1}\ |$are the voltage magnitudes at buses $i$and $k$, respectively; $\theta_{ik,t+1} = \theta_{i,t+1} - \theta_{k,t+1}$is the voltage angle difference between buses $i$and $k$; $G_{ik}$and $B_{ik}$are the conductance and susceptance entries of the bus admittance matrix $Y_{ik}$; and the summation is carried out over all buses $k = 1, \ldots, 6$.

From this, the complex voltages $V_{i,t+1} = |V_{i,t+1}|\angle\theta_{i,I+1}$ and the active and reactive power injections of the slack generator are extracted. Then, the direction of branch currents $I_{ij,t+1}$ and their apparent power flow $S_{ij,t+1,l}$ are obtained from:

$$\begin{pmatrix} I_{ij,t+1} \\ I_{ji,t+1} \end{pmatrix} = \begin{pmatrix} \frac{1}{|t_{ij}|^2}\left(y_{ij} + y_{ij}^{(sh)}\right) & -\frac{1}{|t_{ij}^*|^2} y_{ij} \\ -\frac{1}{|t_{i,j}|^2} y_{ij,} & \left(y_{ij} + y_{ij}^{(sh)}\right) \end{pmatrix} \begin{pmatrix} V_{i,t+1} \\ V_{j,t+1} \end{pmatrix} \ldots\ldots..13$$

$$|S_{ij,t+1,l}| = |V_{i,t+1} I_{ij,t+1}^*| \ldots\ldots\ldots\ldots\ldots\ldots\ldots\ldots\ldots\ldots\ldots\ldots\ldots\ldots\ldots\ldots..14$$

$$|S_{ji,t+1}^*| = |V_{j,t+1} I_{ji,t+1}^*| \ldots\ldots\ldots\ldots\ldots\ldots\ldots\ldots\ldots\ldots\ldots\ldots\ldots\ldots\ldots\ldots 15$$

#### 4.1.2. Reward function

The reward function proposed in the ANM6-Easy environment is formulated based on a multi-objective criterion, and mainly centering on the minimization of equipment energy losses as well as slightly minimizing renewable energy curtailments:

$$r_{f_i} = -\left(\Delta E_{t:t+1} + \lambda\Phi(\mathrm{s}_{f+1})\right) \ldots\ldots\ldots\ldots\ldots.16$$

The total energy loss term is broken down into three separate sub-terms, corresponding to the main sources of inefficiency in this smart grid system architecture

Losses on transmission: They are the energy that is wasted by leaks and resistance in substations and lines of transmission.

DES operational losses This takes into account the net energy that reaches a certain part of the Distributed Energy Storage (DES) unit. It is an approximation to the losses caused by battery charging and discharging inefficiencies, see [15].

$$^{2}\theta_{1k.} = \theta_{i,} - \theta_{k,.} \dots\dots\dots\dots 17$$

Curtailments Renewable Energy: These losses occur when the entire power output from renewable sources cannot be accommodated in the grid. This is usually due to either lack of demand or line

In mathematical terms:

$$\Delta E_{t:I+1.} = \Delta E_{t:t+1.}^{(1)} + \Delta E_{t:t+1}^{(2)} + \Delta E_{t:t+1}^{(3)} \dots\dots\dots\dots\dots..18$$

$$\Delta E_{t:t+1.}^{(1)} = \Delta t \sum_{d\in\mathcal{D}.} P_{d,t+1} \dots\dots\dots\dots\dots\dots\dots\dots\dots\dots\dots\dots 19$$

$$\Delta E_{i:t+1.}^{(2)} = -\Delta t P_{\mathrm{DES},t+1} \dots\dots\dots\dots\dots\dots\dots\dots\dots\dots\dots..20$$

$$\Delta E_{t:t+1}^{(3)} = \Delta t \sum_{g\in D_g - \{g^{(\mathrm{slack})}\}.} \left(P_{g,t+1}^{(\max)} - P_{g,I+1}\right) \dots\dots 21$$

The penalty function includes two particular grid related constraints; the rated power of lines and substations and the bus voltage limits. These restrictions are imposed to avoid the overheating of equipment and the voltage stability of devices connected.

$$\Phi(\mathbf{s}_{i+1}) = \Delta t\left(\sum_{i=1}^{6} \left(|V_{i,t+1}| - \bar{V}_{i,}\right)^{+} + \left(\underline{V}_{i,} - |V_{i,t+1}|\right)^{+} + \sum_{i,j} \left(|S_{ij,t+1}| - \bar{S}_{ij}\right)^{+} + \left(\underline{S}_{ij} - |S_{ij,t+1}|\right)^{+}\right)^{+} \dots\dots\dots\dots\dots\dots.22$$

In our situation $\lambda = 100$ and, for numerical constancy, $r_i$ is clipped to the range $[-100,100]$.

#### 4.1.3. Terminal state

It is considered as a terminal state if there is no physically realizable le solution to equations (9)-(10) under selected actions. This phenomenon corresponds to grid failure, which is usually induced by the violation of voltage constraints. This is called a sag which is one of the major power quality disturbances [51] with important negative effects, especially in industry.

### 4.2. Surrogate model

The surrogate environment for ANM6-Easy is designed to emulate state evolution in a Markov Decision Process (MDP). It is trained to forecast the next system state and the reward of both states:

$$(s_{t+1}, r_t) = \text{Surrogate}(s_t, a_t) \dots\dots\dots\dots..22$$

The main role of this surrogate model is to speed up calculation; the transition speeds of such a surrogate model can be orders of magnitude faster than those recorded by the original simulation. The architecture consists of three specialized neural network modules for the power transition dynamics of non-slack generators, battery systems and the global power balance, respectively. Although power systems are highly interconnected entities characterized by non-linear descriptions, this structure preserves physical interpretation as an inferential procedure that depends on the sequential order of data rather than isolated parallel updates.

However, our architecture is explicitly designed to maintain this physical coupling through a sequentially dependent inference process. The models do not operate in parallel isolation during the prediction of a state transition. Instead, they form a computational chain that respects the cause-and-effect relationships of the grid physics (see reference [45] for more details on equations in section 4.2)

A Model for State Transition The simulation of a single state transition from $s_t$to $s_{t+1}$is implemented in the following stages:

1. Asset Power Forecasting: Two KKT-Informed Neural Networks [45] are first used to estimate the active and reactive power production of the non- slack generators, and battery respectively. These predictions are premised on control actions $a_t$ and state variables from $s_t$,, like the State of Charge (SoC). These networks are trained with a loss, which is KKT-informed in order to guarantee a strict fulfillment of the local operating constraints.
2. Aggregation To achieve this, the forecast power injections of the devices as well as all individual nodes are aggregated for each bus in the network. These output bus powers, $\left\{P_{t+1}^{(i)}\right\}_{i=2}^{6}$and $\left\{Q_{t+1}^{(i)}\right\}_{i=2}^{6}$, are the direct outputs of the final module.
3. Power Balance and Voltage Prediction: The aggregated injections are processed by the power balance network to predict (the system wise states, 1) bus voltage $\{|V_i|, \theta_i\}_{i=2}^{6}$ and 2) the power consumption at the slack generator. This block is learned to approximate the nonlinear power flow equations by reducing the residuals of grid's physical laws.

This cascade connection allows the outputs of the asset models to directly influence the inputs of system-wide power balance model causally. By adding coupled modules, the physical connection of the grid is explicitly imposed in order that the surrogate can adequately reproduce how the network reacts to certain power injections.

### 4.2.1. Generators state

As explained in section 4.1.1, transformation ( $a_{P_{g,}}, a_{Q_{g,}}$ ), to $\left(P_{g,t+1}, Q_{g,t+1}\right)$ is performed by solving the convex optimization problem formulated by (5). The problem presents in a standard form :

$$\min_{\left(P_{g,t+1}, Q_{g,t+1}\right)} \left\| \left(a_{P_{g,t}}, a_{Q_{g,t}}\right) - \left(P_{g,t+1}, Q_{g,t+1}\right) \right\| \quad \text{s.t. } G\left(P_{g,t+1}, Q_{g,t+1}\right) - h \leq 0 \quad \ldots\ldots\ldots\ldots\ldots 23$$

$$G = \begin{pmatrix} -1 & 1 & 1 & 0 & 0 & -\tau_g^{(1)} & \tau_g^{(2)} \\ 0 & 0 & 0 & -1 & 1 & 1 & 1 \end{pmatrix}^T \ldots\ldots\ldots\ldots\ldots\ldots 24$$

$$h = \begin{pmatrix} \underline{P}_g & \bar{P}_g & P_{g,t}^{(\max)} & \underline{Q}_g & \bar{Q}_g & \rho_g^{(1)} & -\rho_g^{(2)} \end{pmatrix}^T \ldots\ldots\ldots\ldots 25$$

Since the formulation in (21) is a convex program with Slater's condition $\left(P_{g,t+1}, Q_{g,t+1}\right)^*$ holding for its constraints, an optimal solution exists if and only if there exist dual variables $\lambda^* \in \mathbb{R}^7$ such that the following Karush–Kuhn–Tucker (KKT) conditions hold.

We use a KKT-Informed Neural Network to predict $\left\{\left(P_{g,t+1}, \underline{Q_{g,t+1}}\right)^*\right\}_{g \in \mathcal{D}_\delta - (g_{\text{glack}},4}$ from the given inputs. Contrary to conventional data-driven models, this approach casts the KKT-optimality conditions as part of the loss function itself.

$$\left\{a_{P_{g,t}}, a_{Q_{g,t}}, P_{g,t}^{(\max)}\right\}_{g \in \mathcal{D}_G - (g^{\text{slack}},\}} \ldots\ldots\ldots\ldots\ldots\ldots 26$$

### 4.2.2. Battery state

Likewise, the battery system operational constraints as stated in (7) are rewritten into the standard optimization form as:

$$\min_{(P_{\text{DES},t+1},Q_{\text{DES},t+1})} \left\| \left(a_{P_{\text{DES},t}}, a_{Q_{\text{DES},t}}\right) - \left(P_{\text{DES},t+1}, Q_{\text{DES},t+1}\right) \right\| \quad \ldots 27$$

$$\text{s.t. } G\left(P_{\text{DES},t+1}, Q_{\text{DES},t+1}\right) - h \leq 0$$

$$G = \begin{pmatrix} -1 & 1 & 0 & 0 & -\tau_{\text{DES}}^{(1)} & \tau_{\text{DES}}^{(2)} & \tau_{\text{DES}}^{(3)} & \tau_{\text{DES}}^{(4)} & -1 \\ 0 & 0 & -1 & 1 & 1 & 1 & 1 & 1 & 0 \\ & & & & & & & & 0 \end{pmatrix}^T \quad \ldots 28$$

$$h = \begin{pmatrix} \underline{P}_{\text{DES},} & \bar{P}_{\text{DES}} & \underline{Q}_{\text{DES},} & & \bar{Q}_{\text{DES}} \\ & \rho_{\text{DES}}^{(1)} & -\rho_{\text{DES}}^{(2)} & -\rho_{\text{DES}}^{(3)} & \rho_{\text{DES}}^{(4)} \\ & & & & \frac{1}{\eta \Delta t}\left(\text{SoC}_t - \overline{\text{SoC}}\right), \frac{\eta}{\Delta t}\left(\text{SoC}_t - \underline{\text{SoC}}\right) \end{pmatrix}^T . \quad \ldots 29$$

So a KKT-Informed Neural Network is trained to predict the optimal active and reactive power $\left(P_{\text{DES},t+1}, Q_{\text{DES},t+1}\right)^*$ using $\left[a_{P_{\text{DES},t}}, a_{Q_{\text{DES},t}}, \text{SOC}_t\right]$ as the input vector. By incorporating the optimality conditions of the battery's feasible region and its dynamic limits (e.g. SoC), such a network forces the predicted charge/discharge dynamics to be consistent with those imposed by hardware design.

### 4.2.3. Power balance

After the active and reactive power of each device (except for the slack generator) are solved, the results are gathered at all their corresponding buses to get coarse-grained bus-wise quantities $P_i^{(\text{bus})}$ $Q_i^{(\text{bus})}$ $i = 2, \ldots, 6$

The power balance network takes these accumulated powers as input to forecast the bus voltages $\{|V_i|, \theta_i\}_{i=z^-}^{6}$. Since the slack bus voltage is maintained fixed ( $V_1 = 1$ p.u. and $\theta_1 = 0°$ ), one can calculate the residuals of power flow equations (9-10) for all other buses for $i = 2, \ldots, 6$. The network learns to minimize the difference between the predictions and ground truth by minimizing the following loss per sample:

$$\mathcal{L} = \mathcal{L}_1 + \mathcal{L}_2 \quad \ldots\ldots\ldots\ldots\ldots\ldots\ldots\ldots\ldots\ldots\ldots\ldots.30$$

$$\mathcal{L}_1 = \frac{1}{5}\sum_{i=2}^{6}\left(P_i^{(\text{bus})} - \sum_{k=1}^{6} |V_i||V_k|(G_{ik}\cos\,\theta_{ik} + B_{ik}\sin\,\theta_{ik})\right)^2 \quad \ldots\ldots\ldots 31$$

$$\mathcal{L}_2 = \frac{1}{5}\sum_{i=2}^{6}\left(Q_i^{(\text{bus})} - \sum_{k=1}^{6} |V_i||V_k|(G_{ik}\sin\,\theta_{ik} + B_{ik}\cos\,\theta_{ik})\right)^2 - \cdots \ldots\ldots\ldots.32$$

After obtaining $\{|V_i|, \theta_i\}_{i=1}^{6}$ the active and reactive power injection of the slack generator are obtained as:

$$P_1^{(\text{bus})} = \sum_{k=1}^{6} |V_1||V_k|\big(G_{1,k}\cos\,\theta_{1k.} + B_{1k}\sin\,\theta_{1k}\big) \ldots\ldots\ldots\ldots\ldots.33$$

$$Q_1^{(\text{bus})} = \sum_{k=1}^{6} |V_1||V_k|(G_{1k.}\sin\,\theta_{1k.} + B_{1k}\cos\,\theta_{1k}). \ldots\ldots\ldots\ldots\ldots.34$$

### 4.2.4. Next state

Assuming perfect foresight for the future of load and generation, we treat$\{P_{d,t+1}\}_{d\in D_{L.}}$ and $\{P_{g,I+1}^{(\max)}\}_{g\in D_G-\{g^{\text{slax}}\},}$ as known. Now, equation (3) and (8), with the above networks created in sections 4.2.1 - 4.2.3, result in:

$$\text{aux}_{t+1} = (\text{aux}_t + 1)\text{mod}96 \ldots\ldots\ldots\ldots\ldots\ldots.35$$

There is already enough information in all these defects, to build the next state $\mathbf{s}_{t+1}$.

### 4.2.5. Reward

All information to compute (11)-(19) is available.

## 5. Results and discussion

### 5.1. Surrogate training conditions

Each of the three constituent modules of the surrogate is defined as a multilayer perceptron with three hidden layers of 512 neurons each, a residual connection between the inputs and outputs of the latter, and a Leaky RELU (slope of 0.01) as the activation function.

Each module is optimized with Adam W, with a learning rate of $1 \times 10^{-5}$. An early stopping condition is imposed: training stops when no progress is observed in the last 5000 steps.

At each training step, a batch of 64 samples is sampled from a Sobol sequence of dimensionality 21. All dimensions are scaled to provide inputs for each part of the surrogate, as outlined in Section 4.2. For more information on the scaling process,

Our proposed PINN-based surrogate model has been analyzed against other methods from the state of the art, considering seven types of surrogates trained from two different kinds of datasets. The studied surrogates are: Deep Neural Networks (DNN), XGBoost (XGB), Decision Trees (DT), Random Forest (RF), Linear Regression (LR), and our proposed method, with PINNs.

These models (except for the PINN) were trained using two types of datasets: Generative and Agent-based. Following the naming convention established in [44], a Generative dataset is constructed by sampling transitions across the state space using a general sampling method, in this case Sobol sampling [52]. In contrast, an Agent-based dataset is derived from the agent interaction of a specifically designed agent with the environment, represented here by a Random agent. Both datasets consist of 100,000 samples.

The loss function used in the training process is the Mean Absolute Error (MAE) for each model. However, the coefficient of determination metric ($R^2$) has been used to evaluate the performance of the trained models.

The results reported in the following sections have been obtained by training the models on an Intel Core i5-9499F CPU.

### 5.2. Surrogate training results

This section presents the results regarding the surrogates' training metrics. Figure 4 shows trends of losses for the generators and the DES network, as defined in [45], and for the power balance solver, over the training.

Additionally, the speed increase of the surrogate compared to the original environment is calculated by performing 1000 transitions with both. As shown in Fig. 5, there is a median boost of almost 10 times.

Figure 7 shows the accuracy comparison between all the studied models in terms of $R^2$ and MAE. In line with the conclusions of Cestero et al. [44], these results generally show that agent-based sampling produces better performance than generative sampling in most cases. This is clearly seen in the MAE metric, which is plotted on a logarithmic scale, and the error of the models trained on the generative dataset is generally much higher than that of the other.

XGB, followed closely by the PINN model, from these results, we can expect that XGB and PINN models perform best as a proper environment for RL training, while the models DNN and LR may have the least favorable performance. However, this is not what happens in practice, as further results reveal. Once the models operate directly within the environment, their performance shifts significantly.

The overall accuracy of the models in terms of $R^2$ suggests that many models, in principle, should be able to represent the transitions of the environment accurately. To validate this, we run random episodes using two different agents: a pre-trained agent that is able to follow the optimal policy in the environment, and an expert agent, i.e., an agent that takes actions according to the environment dynamics. We then calculate the MAE of all the surrogates throughout the episodes of each agent. Figure 7 shows the averaged overall MAE of each surrogate model. The results are obtained by running all the surrogates in parallel with the real environment and following the real transitions from the environment, calculating transitions with the surrogates too. Then, the error is calculated as the MAE of the surrogate transitions compared to the real ones, and the MAE is averaged between several episodes.

The error is presented in Fig. 8 as averaged MAE of surrogate transitions across the episodes, calculated as the MAE of the difference between surrogate predictions and real transitions. In addition, there is no accumulated error in this figure, since the surrogates do not simulate the episode independently; they are only used as a transition predictor. To calculate the transition $s_{t+1} = f(s_t, a_t)$, the surrogate functions $f_{\text{surrogate}}$are evaluated at each step, and the real state taken from the real environment is used for the next step.

As per the figure, our method shows the least error, even though Fig. 5 shows that other models display more accurate overall results. These results highlight that PINNs present the most suitable method for modeling datasets with physical extrapolation capabilities, while the PINN model is reliable across the entire state space.

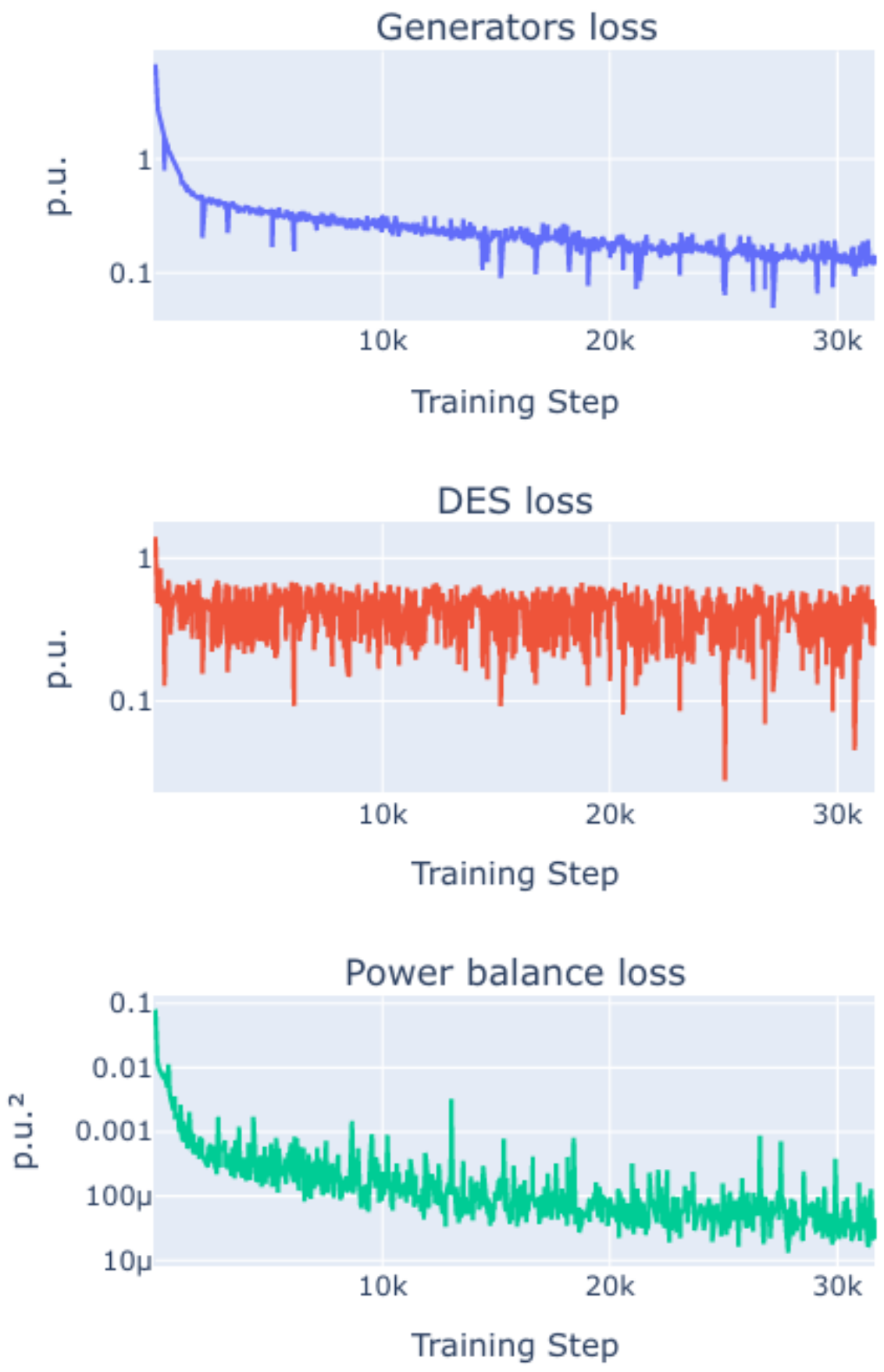

*Figure 5 Loss function components. The training metrics are such that: Generators Loss is a sum of KKT residuals associated with the generator-related KKT-NN, DES Loss is similarly formed by summing up KKT residuals from battery-related KKT-NN and Power Balance Loss is computed as the error in power balance equations (27) coupled with voltage related PINN.*

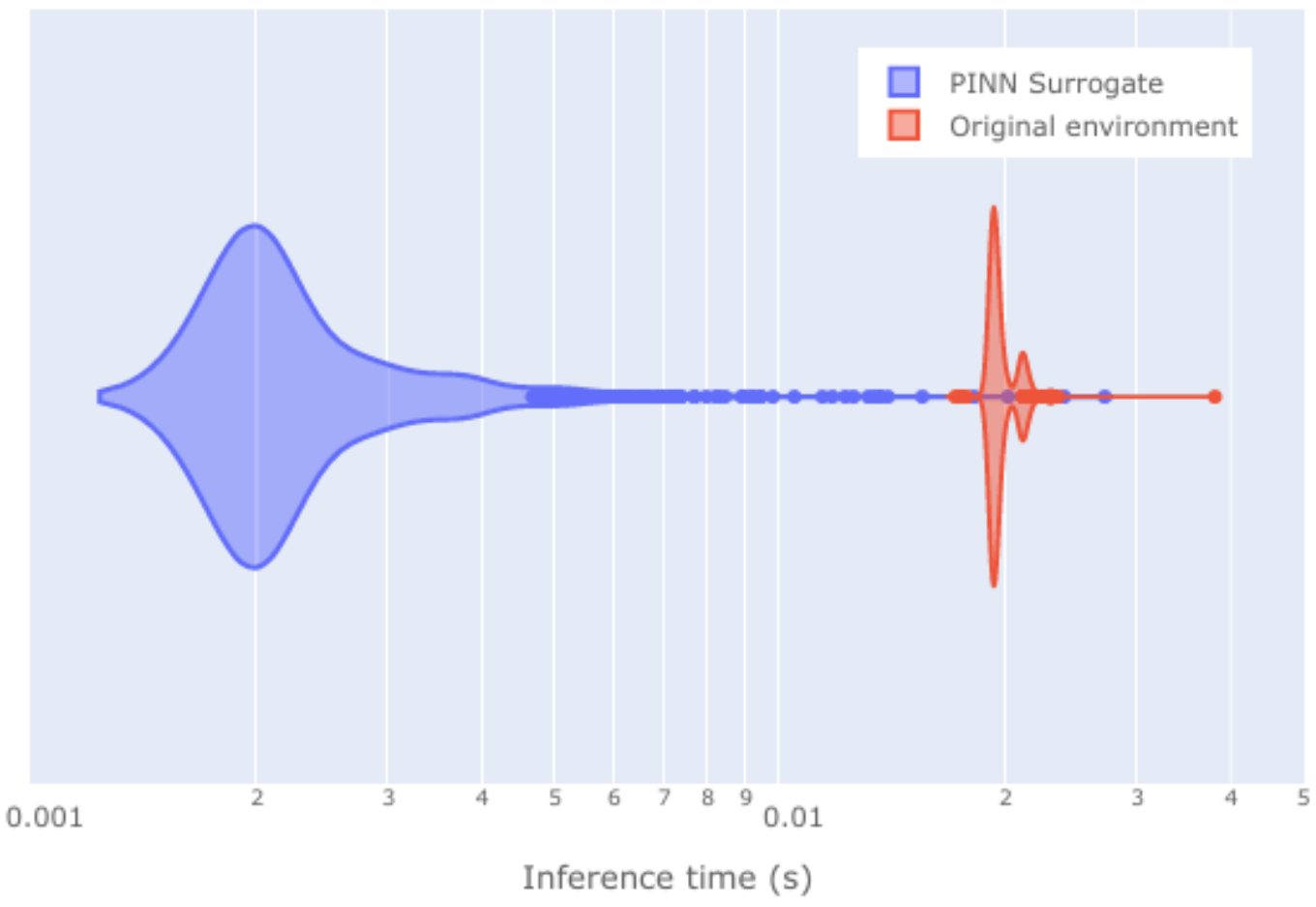

*Figure 6 Analysis of  Inference time (1,000 transition). This is a large difference in  computational cost, since the PINN has a median inference time of 0.0021 s and the plain environment a median one of 0.0194 s.*

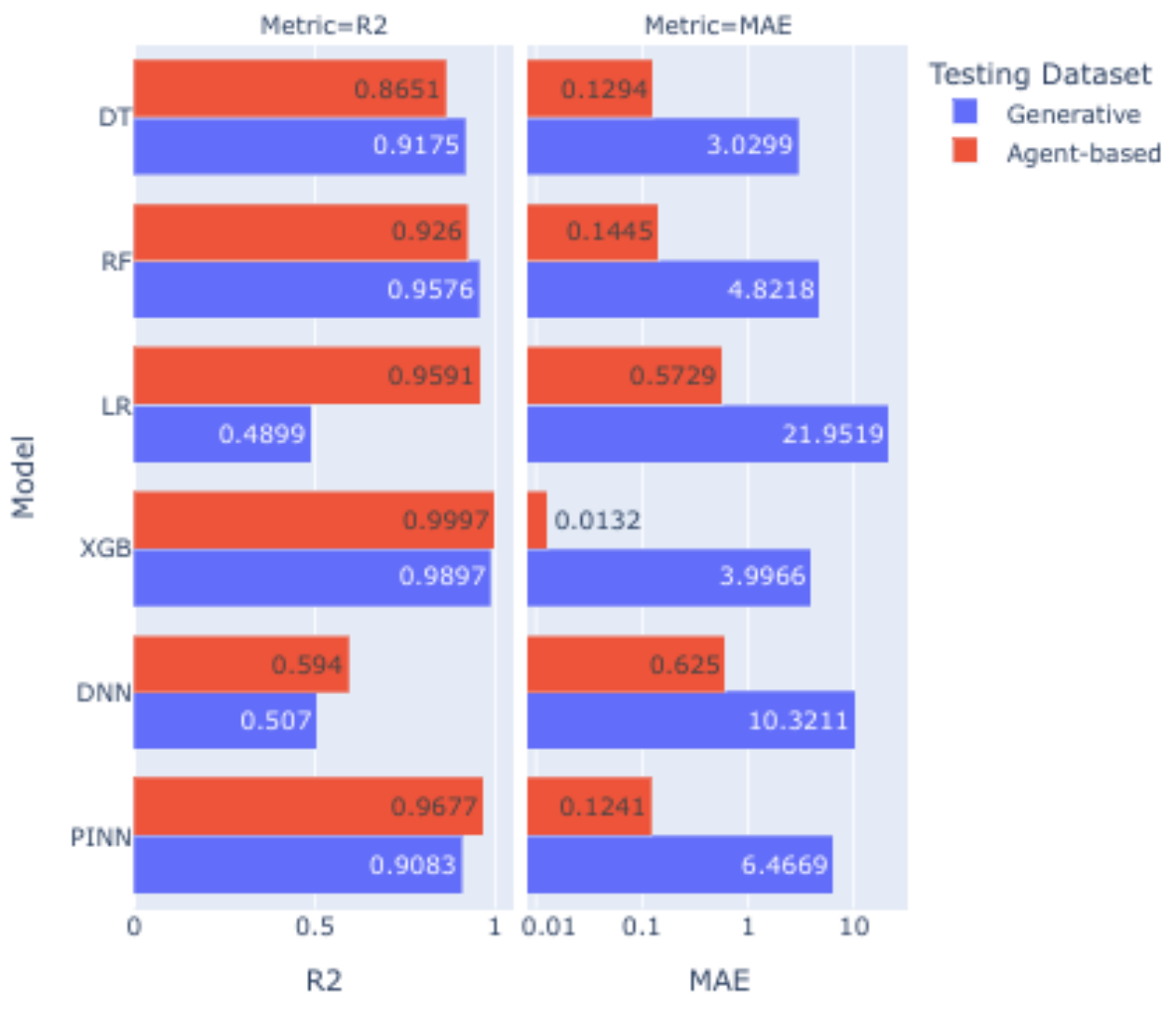

Figure 7 Model performance assessment evaluation of surrogate models with $R^2$ and MAE  metrics. Table 1 Compares the accuracy of different surrogate models, $R^2$ is  shown in linear scale and MAE on log scale(options include Genertive and Agent-based datasets). Although XGB models return the best performance in terms of accuracy which is accompanied by the smallest error—barely outperforming PINN—all except  P NJ b1 and 8 are considered to be incapable for RL training.

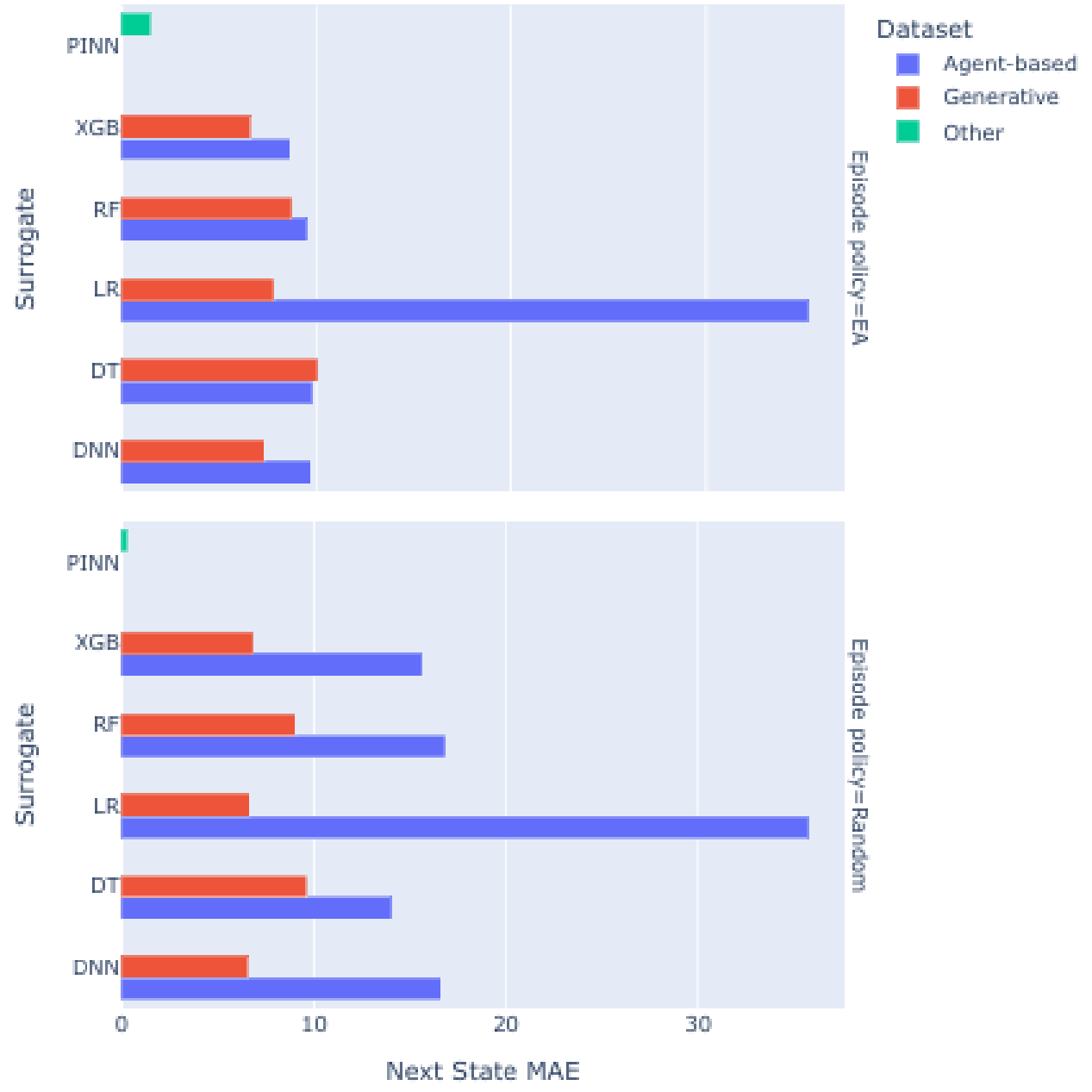

*Figure 8 Distribution of MAE by training dataset. The errors of the various model are computed from training data (Agent-based or GAN) and averaged over a random episode, using both an Expert Agent (EA), as well as Random policy. In all these cases the value of the error is minimized by the model with PINNs. Observethat "Oth r" in dicates using no train d taset for the surrogate.*

Important conclusions about the performance of surrogates trained on Agent-based versus Generative data are provided in this paper. Nevertheless previous work in the group of Cestero et al. [44], the proposed method does not use a mixed strategy (random, expert, and search policies) to create the dataset but instead utilizes only a random policy. In contrast, the Generative dataset (generated using Sobol) provides consistent results for all surrogates and an indication of a more uniform coverage of the state space. In comparison, the Agent-based dataset exhibits large differences across explored regions when it is probed with expert or random policies.

The convergence of the Mean Absolute Error (MAE) for the PINN surrogate in episodes where expert and random policies are used is shown in Fig. 8. While we play representative single episodes, the behavior was consistent across multiple runs. There is low error in the random episodes throughout; distinct features of daily periodicity and two-stage development (initial slope followed by stability) are present. These variations in the error profile are explained as periodic patterns in the grid. The small spikes in the MAE correspond to particular state variables which are under-represented the most often—corresponding to transient energy generation or consumption peaks.

Inspection of a simulation rendering shows that this periodicity is determined by two main factors:

- The battery configuration shows a mean charge that continues to grow linearly in time the long run while oscillating (in daily scale), it eventually saturates at steady state.
- The cycle between the daily peaks of energy consumption, such as the evening peak for charging your car, versus a minimum in demand during the dead of night.

Even with these occasional fluctuations, the PINN surrogate maintains a lower error rate than other algorithms. The discrepancy between the performance of the two agents is due to a difference in regions of state space visited; random agents remain near wellvisited initial states, while expert agents follow directed trajectories into states that are often underrepresented in the training data [44]. As a result, the accuracy of PINN is usually higher for states that are visited by a random agent rather than those traversed by an expert.

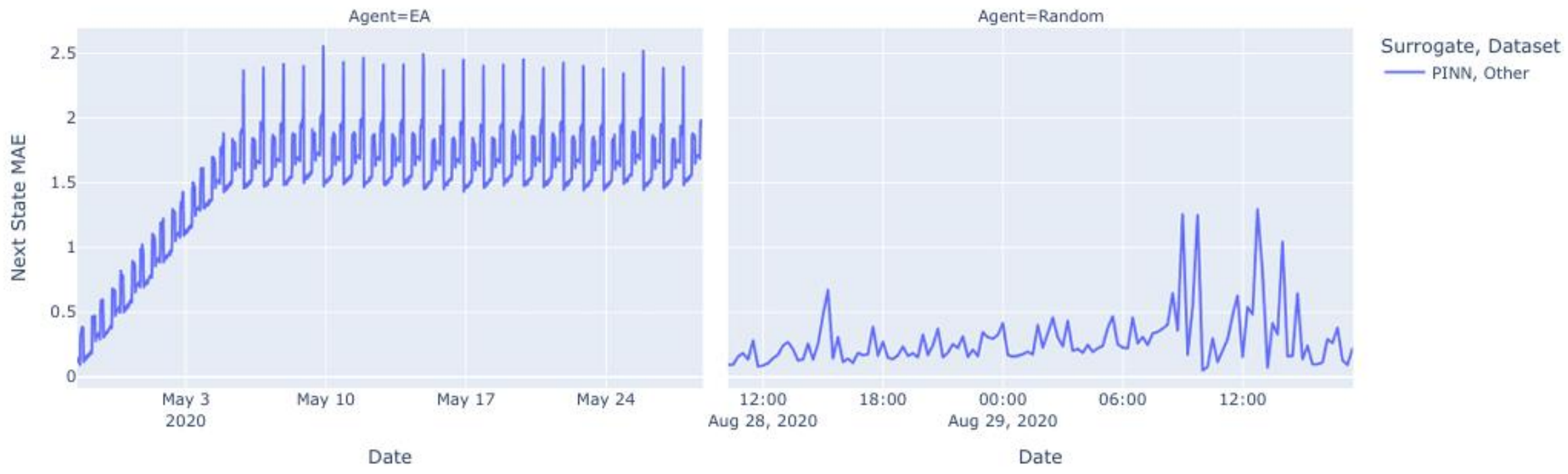

*Figure 9 Progression of the MAE for the PINN surrogate. Error evolution of the PINN agent is studied for a sample episode which takes place with both type of agents -one EA and one RA. It is noted that EA results exhibit a unique 24 daily periodicity in error and a linear increase of an average error followed by plateau. This tendency is mostly due to environmental environment, when stability will be achieved while load and battery will take the "best ones". On the contrary, the Random agent has a small overall error magnitude but with no clear periodicity.*

## 5.3. Optimization of environment structural parameters

The parallelizable nature of surrogate models is leveraged by predicting multiple transitions at once, enabling to grid multiple environment into the same operation. However, it is reported that the use of a large number of parallel environments can lead to training speed saturation and performance decay for PPO from Stable-baselines3. This impairment takes place in the network training phase, when we have the model with an increasing number of samples and thus it gets easily saturated and asymmetrically polarized. As a result, the algorithm’s rollout buffer size is said to be a key parameter affecting the training  speed. It is worth noting that reducing the rollout buffer can make updates faster but also unstable if smaller than an episode duration.

An investigation has been performed for various  options of the numbers of these two variables to search out the optimal combination, and their illustration is provided in Fig. 9. A single episode in this environment has 3,000 steps, and smaller  buffer sizes do not finish a full episode without updating. However, the data clearly indicates that a full episode is not needed for  good policy learning; the best configuration of structural parameters turns out to be 100 parallel environments and a buffer size of 30. Early Termination: Similar early stopping techniques are also employed to terminate  experiments at a performance plateau or after ax number of steps.

A more in-depth analysis of these structural parameters includes computing their relationship to the mean reward per  episode and total training time, as shown in Fig. 10. There is a strong negative correlation between  the mean reward and the buffer size, which are the same effects observed in our early attempts on parallelization. Moreover, the mild  impact of environment number on reward is less significant than that of buffer size. As for the buffers, we find that their total  training time grows proportionally with both buffer size and number of environments, with them having roughly equal impact on the length of time for training.

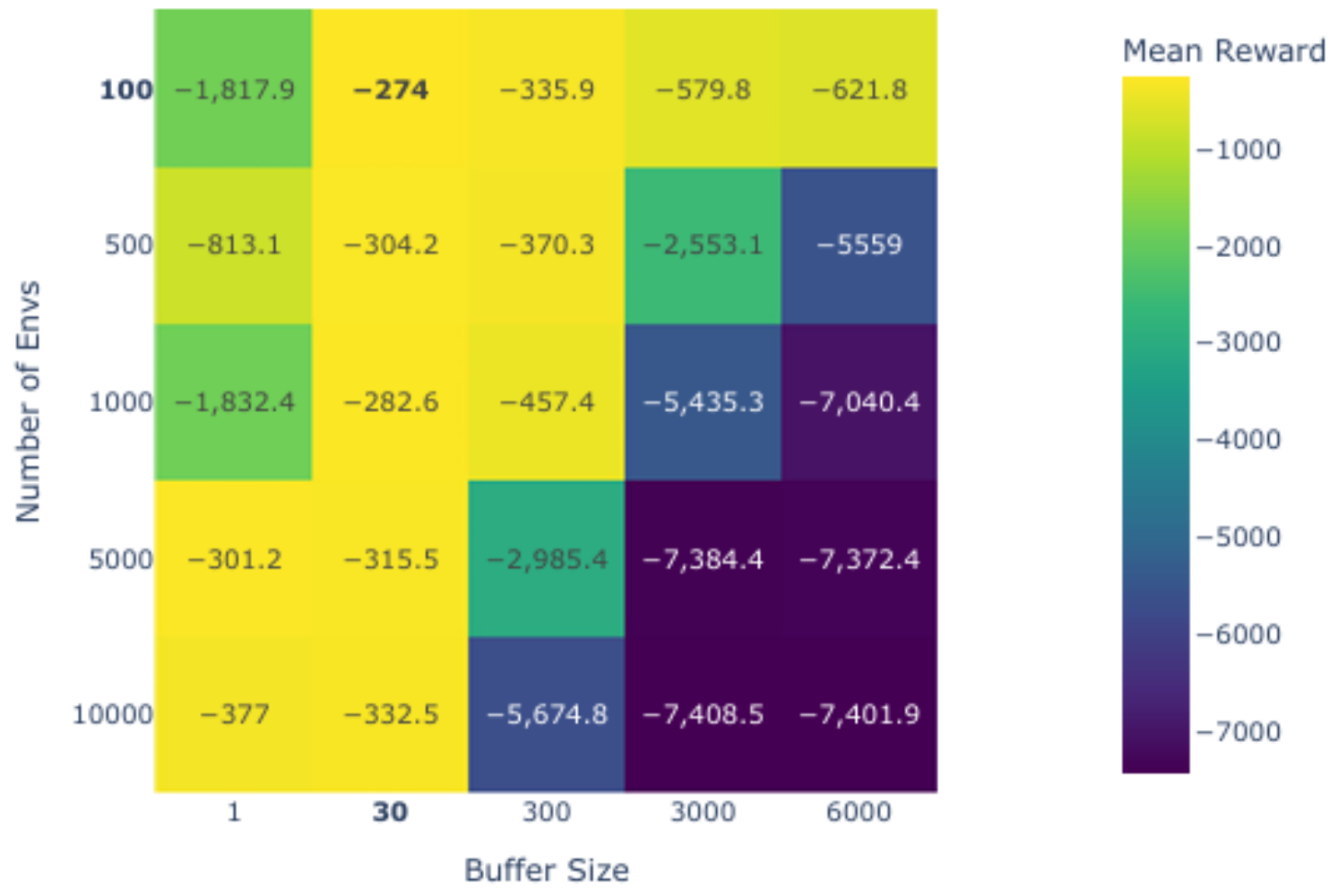

*Figure 10 Results  of the reward acquisition optimization for the structure parameter. The shown results are the  mean reward of each parameter combination over the last ten evaluations. For the sake of consistency, some trials were stopped when they arrived at a maximum step counter or when  convergence was obtained. Note that effective rollout buffer size will be the product of the  number of envs and the hyperparam in this paper.*

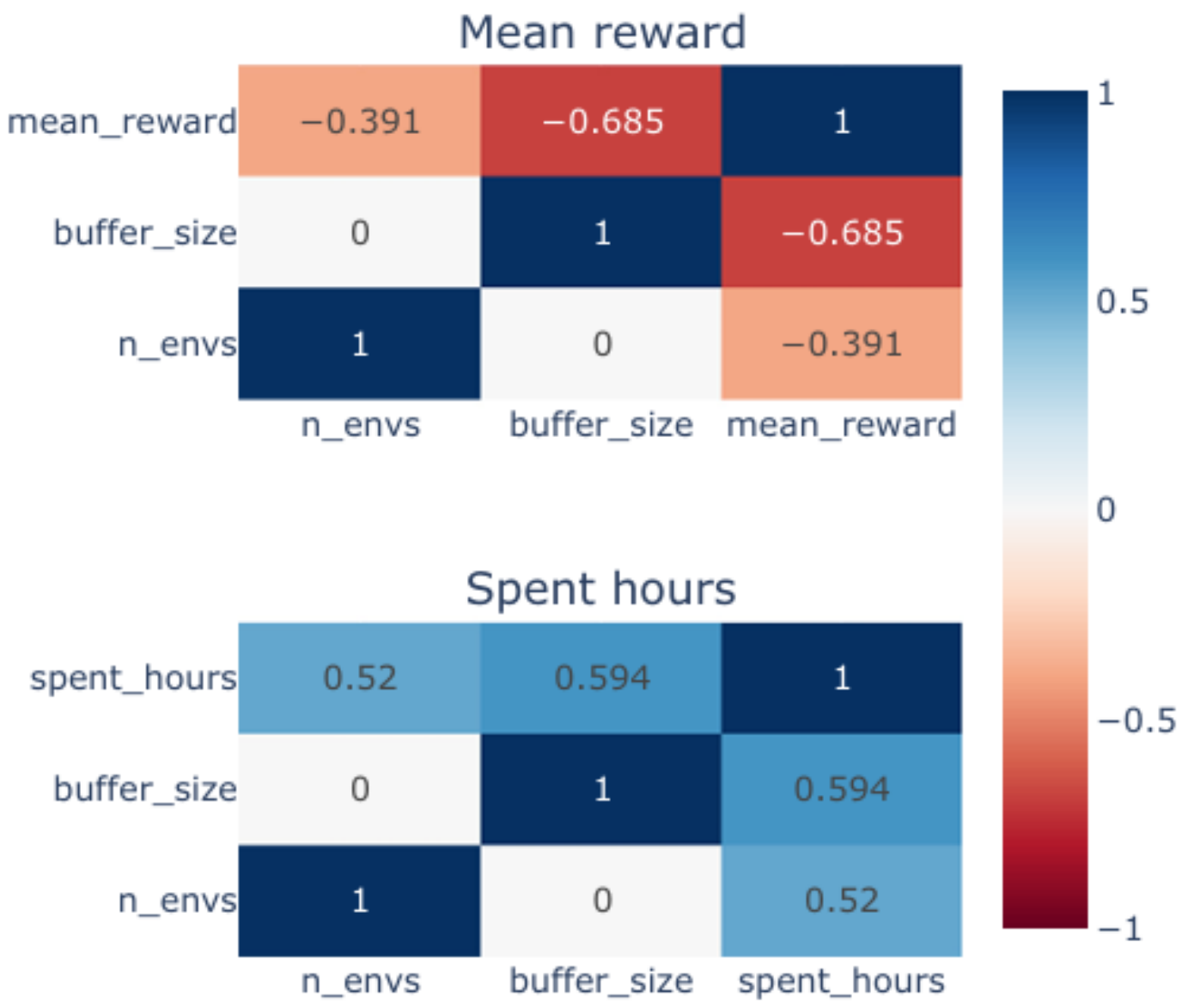

*Figure 11 Correlation analysis of structural parameters. The links among buffer size, n_envs, mean reward and duration of training are visualized. A negative correlation is detected among the reward, the buffer size and the number of environments, where impact of buffer size is greater. Furthermore, the two parameters present a similar positive relationship with total training time.*

With the previously determined optimal structural parameters, each of the investigated models was used as a surrogate habitat to train a PPO agent on an optimal policy. Fig. 11 shows the evolution of the training reward of multiple surrogates compared to a baseline (i.e., training a PPO agent on the original environment with no surrogate). For efficiency, an early stopping condition was added on all agents which stopped the experiment when rewards began to saturate (when the reward had not been improved upon for 20 episodes).

The results show that our method with PINN-trained surrogate is much faster than the baseline with a comparable performance. It is worth pointing out that the PINN surrogate is found to be the only model (when trained with policy optimization) able to effectively generate a primitive policy which achieves a performance gain over random action. That this model can capture the background physics of the environment, and not just mimic what is observed, explains this success.

In terms of operation efficiency, the smart grid controlled by the PINN agent is able to reduce and achieve much faster and reliable reductions in energy losses than all other models including the baseline. Examination of simulation renders shows that the agent correctly recognises and prevents the main sources for power loss. To be more precise, the battery is regularly used between 20% state of charge (SOC) and 80% SOC because these two boundary points are where you lose less energy and put less wear on your battery.

Additionally, the agent shows good knowledge about physical limitations. For example, in the illustrated installation, power supply line to the bus from battery and EV chargers are rated for a capacity less than or equal to peak of EV demand. Even so, the agent is able to use the battery to fill in for power that the line, from a physical standpoint, cannot deliver. Yet another is the agent's extreme disinclination to renounce a considerable amount of renewable energy, even at the expense of diverting power from the slack generator and increasing some losses. This may indicate that the agent has found a sweet spot between voltage penalties and power losses, showing that the reward function effectively values grid safety and equipment lifespan more than consumption of as much renewable energy as possible.

Although Fig. 6 one might also argue that the XGB model would perform best because of its almost perfect $R^2$ value but as we see in contrast to RF, prediction errors increase during long-term simulations with XGB. This causes the RL agent to fail to learn a stable policy. On the contrary, for the PINN surrogates, this problem can be mitigated by learning state transitions deeply and enforcing physical laws strictly so that physical predictions can remain realistic even in such regions of the state space not being visited (see Fig. 7.

On a practical note, the extrapolation ability of PINN surrogate enables the manipulation of demand and generation profiles while no retraining is necessary. In practice, consumption on distribution networks often changes due to new loads or tenants and generators can increase their rated generation. The ability to obtain reliable simulations under such new conditions without additional training time or computational burden is a clear advantage of the PINN approach.

**5.4. Discussion**

The findings given in Section 5.3 show that the PINN-based surrogates perform for reinforcement learning policies similar to the training in the original environment, reducing computational time considerably. Instead of elaborating on the error dynamics seen on Expert Agent and Random Agent trajectories, the discussion will look to the potential benefits of these findings for surrogate-assisted reinforcement learning to be used in smart grid control. A crucial observation is that surrogate accuracy is evaluated through traditional regression metrics, like $R^2$or absolute mean error computed on static data sets, is often not sufficient for successful policy learning. Several data-driven surrogates have good predictive strength as a function of working offline and still suffer from the problem of the fact that errors accumulate (accumulation) over long horizon paths in reinforcement learning. This behavior highlights a critical limitation of purely data-driven models: small local inaccuracies can compound over time when the model is repeatedly queried within a closed-loop control setting. By contrast, the PINN-based surrogate maintains physically consistent state transitions even when extrapolating beyond densely sampled regions of the state space. By explicitly embedding power flow equations and device-level operational constraints into the learning process, the surrogate keeps feasibility and stability properties which are important for reinforcement learning. This may explain why the PINN surrogate works well when the policy goes under scenarios that the training data does not have very much coverage of, while pure data surrogate departs from this paradigm only after similar conditions. The learned policies demonstrate behavior matching established engineers' practices from a control perspective. Given the endurance of the battery within intermediate state-of-charge ranges and prioritizing grid safety above aggressive renewable utilization, it can say that the reward formulation along with the physics-informed surrogate provides the means for physically meaningful control strategies. Significantly, these behaviors arise without the explicit hard-coding, indicating that the framework is capable of learning from the system constraints and internalizing them. Another implication is adaptability. Since the PINN surrogate encodes governing physical relationships rather than historical paths, it does not need to be retrained in order to support changes of demand profiles or generation patterns, unlike its ancestors. This is not only useful for real-world distribution networks, where operating conditions change over time due to new load factors, generation assets or user behavior. In conclusion, in the present study, we assume a deterministic forecasting to separate the impact of the surrogate model, but our framework will apply to stochastic settings, automatically by introducing forecasting uncertainty in the state representation. These results imply that physics-guided alternative scenarios can provide a strong platform for scalable, data-efficient reinforcement learning in complex energy systems.

## Conclusion

In this work, we develop the capability of Physics-Informed Neural Networks (PINNs) as surrogate models for effective energy management in Smart Grids based on Reinforcement Learning (RL). We developed a Smart Grid simulation using the Gym-ANM framework and trained an RL agent for grid adaptation. The proposed PINN-based surrogate method significantly outperforms traditional approaches to achieve improved computational efficiency and a 10× speed-up in inference time and 50% reduction in policy training time, preserving policy accuracy and reliability. Classical surrogate models did not converge, whereas the PINN surrogate provides stable, efficient learning to cope with the increasing complexity in modern Smart Grids, especially given the high penetration of renewable energy and dynamic consumption. These results exhibit obvious benefits for the optimization of Smart Grid in that their training and inference time is much shorter with very low error which reduces computational cost. The PINN surrogate accurately captures the dynamics of the grid while dynamically changing the demand and generation, thus making it safe and effective to operate at lower maintenance costs and improved system reliability. Hence, this study demonstrates the usefulness of PINN-based surrogate models for speeding up the development of efficient, cost-effective and sustainable management strategy of Smart Grid with real-time operation. These models can be applied to larger systems for the scalability of the approach (e.g., on demand response and energy pricing scenarios for more extensive assessment) to evaluate their performance in real-world uncertain conditions. Adding forecasting models and measuring policy generalization will be key steps toward field implementation.

**Appendix A. Ranges of inputs for surrogate training**

Refer to [15] for a description of the physical parameters that outline the subsequent ranges.

$$a_{P_{g,t}} \in [\underline{P}_g, \bar{P}_g]^{64} \ \forall g \in \mathcal{D}_g$$
$$a_{Q_{g,t}} \in [\underline{Q}_g, \bar{Q}_g]^{64} \ \forall g \in \mathcal{D}_g$$
$$a_{P_{\mathrm{DES},t}} \in [\underline{P}_{\mathrm{DES}}, \bar{P}_{\mathrm{DES}}]^{64}$$
$$a_{Q_{\mathrm{DES},t}} \in [\underline{Q}_{\mathrm{DES}}, \bar{Q}_{\mathrm{DES}}]^{64}$$
$$P_{g,t}^{(\max)} \in [\underline{P}_g, \bar{P}_g]^{64} \ \forall g \in \mathcal{D}_g$$
$$P_i^{(\mathrm{bus})} \in [\underline{P}_i^{(\mathrm{bus})}, \bar{P}_i^{(\mathrm{bus})}]^{64} \ i = 1, \ldots, 6$$
$$Q_i^{(\mathrm{bus})} \in [\underline{Q}_i^{(\mathrm{bus})}, \bar{Q}_i^{(\mathrm{bus})}]^{64} \ i = 2, \ldots, 6$$
$$\mathrm{SoC} \in [\underline{\mathrm{SoC}}, \overline{\mathrm{SoC}}]^{64} \ i = 2, \ldots, 6$$